\documentclass{article} 
\usepackage{nips13submit_e,times}
\usepackage{hyperref}
\usepackage{url}
\usepackage{graphicx}

\newcommand{\bfx}{\mathbf{x}}

\newcommand{\bfw}{\mathbf{w}}
\newcommand{\Xcal}{\mathcal{X}}

\newcommand{\Fcal}{\mathcal{F}}

\newcommand{\Scal}{\mathcal{S}}
\newcommand{\E}{\mathbb{E}}

\usepackage{graphicx} 
\usepackage{subfigure} 

\usepackage{amssymb}
\usepackage{amsmath}

\usepackage{algorithm}
\usepackage{algorithmic}

\newtheorem{theorem}{Theorem}[section]
\newtheorem{proposition}[theorem]{Proposition}
\newtheorem{definition}[theorem]{Definition}

\newcommand{\RR}{\mathbb{R}}

\title{Learning Non-Linear Feature Maps}

\author{
Dimitrios Athanasakis\thanks{ Use footnote for providing further information
about author (webpage, alternative address)---\emph{not} for acknowledging
funding agencies.} \\
Department of Computer Science\\
University College London\\
London, UK \\
\texttt{dathanasakis@cs.ucl.ac.uk} \\
\And
John Shawe-Taylor \\
University College London \\
Address \\
\texttt{jst@cs.ucl.ac.uk} \\
\AND
Delmiro Fernandez-Reyes \\
National Institute for Medical Research \\
Address \\
\texttt{dfernan@nimr.mrc.ac.uk} \\
}

\nipsfinalcopy 

\begin{document}

\maketitle

\begin{abstract}

Feature selection plays a pivotal role in learning, particularly in areas were parsimonious features can provide insight into the underlying process, such as biology. Recent approaches for non-linear feature selection employing greedy optimisation of Centred Kernel Target Alignment(KTA), while exhibiting strong results in terms of generalisation accuracy and sparsity, can become computationally prohibitive for high-dimensional datasets. We propose randSel, a randomised feature selection algorithm, with attractive scaling properties. Our theoretical analysis of randSel provides strong probabilistic guarantees for the correct identification of relevant features. Experimental results on real and artificial data, show that the method successfully identifies effective features, performing better than a number of competitive approaches.

\end{abstract}


\begin{verbatim}
  Feature Selection, Kernels
\end{verbatim}

\section{Introduction}

Feature selection is an important aspect in the implementation of machine learning methods. The appropriate selection of informative features can reduce generalisation error as well as the storage and processing requirements for large datasets. In addition,  parsimonious models can provide valuable insight into the relations underlying elements of the process under examination. There is a wealth of literature on the subject of feature selection when the relationship between variables is linear. Unfortunately when the relation is non-linear feature selection becomes substantially more nuanced. 

Kernel methods excel in modelling non-linear relations. Unsurprisingly, a number of kernel-based feature selection algorithms have been proposed.  Early propositions, such as Recursive Feature Elimination(RFE) [1] can be computationally prohibitive, while attempts to learn a convex combination of low-rank kernels may fail to encapsulate nonlinearities in the underlying relation. Recent approaches using explicit kernel approximations can capture non-linear relations, but increase the storage and computational requirements. The successful use of a kernel-based feature selection methods is a matter of balance. 

\subsection{Related Work}

Our approach makes extensive use of \emph{Kernel Target Alignment} (KTA) [2,3]. Work in [4] provides the foundation of using the alignment of centred kernel matrices as the basis for measuring statistical dependence. The Hilbert-Schmidt Independence criterion is the basis for further work in  [5], where greedy optimisation of centred alignment is employed for feature selection. Additionally, [5] identifies numerous connections with other existing feature selection algorithms which can be considered as instances of the framework.

Stability selection [6] is a general framework for variable selection and structure estimation of high dimensional data. The core principle of stability selection is to combine subsampling with a sparse variable selection algorithm.  By repeated estimation over a number of different subsamples, the framework keeps track of the number of times each variable was used, thus maintaining an estimate for the importance of each feature. More importantly, stability selection provides finite sample control for some error rates of false discoveries and hence a principled approach for parameter selection. In this work, we propose a synthesis of the two aforementioned approaches through a randomised feature selection algorithm based on estimating the statistical dependence between bootstrapped random subspaces of the dataset in RKHS. The dependence estimation of random subsets of variables is similar to the approach of [13], which is extended through bootstrapping and carefully controlled feature set sizes.  

This approach is simple to implement and compares favourably with other methods in terms of scalability. The rest of the paper is structured as follows: \emph{Section 2} presents the necessary  background  on feature selection for kernel-based learning. \emph{Section 3} introduces a basic randomised algorithm for nonlinear feature selection, along with some simple examples, while \emph{Section 4} provides some analysis. Extensive experimentation on real and artificial data in \emph{section 5} concludes this paper.

\section{Preliminaries}

We consider the supervised learning problem of modelling the relationship between a $m \times n$ input matrix $X$ and a corresponding $m \times n'$ output matrix $Y$. The simplest instance of such a problem is binary classification where the objective is the learning problem is to learn a function $f: \textbf{x} \rightarrow \textbf{y}$ mapping input vectors $\textbf{x}$ to the desired outputs $\textbf{y}$.
In the binary case we are presented with a $m \times n $ matrix $X$ and a vector of outputs $\textbf{y}$,  $y_i \in \{+1,-1 \}$
Limiting the class of discrimination functions to linear classifiers we wish to find a classifier 
$$f(\textbf{x}) = \sum_i w_i x_i = \langle \textbf{w, x}\rangle$$

The linear learning formulation can be generalised to the nonlinear setting through the use of a  nonlinear feature map $\phi(x)$, leading to the kernelized formulation: 
$$f(x)=\langle w, \phi(x)\rangle = \langle \sum_i a_i y_i \phi(x_i), \phi(x)\rangle = \sum_i a_i y_i k(x_i,x) $$

The key quantities of interest in our approach is the centred kernel target alignment which is defined as:
$$a(C_x,C_y)= { {\langle C_x,C_y \rangle_F}\over{ \|C_x\|_F \|C_y\|_F}} = {{\sum_{i,j}c_{x_{ij}} c_{y_{ij}}}\over { \sum_{i,j}\|c_{x_{ij}}\|\sum_{i,j}\|k_{y_{ij}}\| }  }$$

The matrices $C_x$ and $C_y$ correspond to centred kernels on the features $X$ and outputs $Y$ and are computed as:

$$C=\left[ I-{{11^T}\over m} \right] K \left[ I-{{11^T}\over m} \right]  $$

where $1$, in the above equation denotes the m-dimensional vector with all entries set equal to one.

\section{Development of key ideas}

The approach we will take will be based on the following well-known observation that links kernel target alignment with the degree to which an input space contains a linear projection that correlates with the target.

\begin{proposition}
Let $P$ be a probability distribution on the product space $\Xcal \times \RR$, where $\Xcal$ has a projection $\phi$ into a Hilbert space $\Fcal$ defined by a kernel $\kappa$. 
We have that
\begin{eqnarray*}
\sqrt{\E_{(\bfx,y) \sim P, (\bfx', y')\sim P}[y y' \kappa(\bfx,\bfx')]} =&& \\
&& \hspace{-5cm}= \sup_{\bfw: \|\bfw\| \leq 1} \E_{(\bfx,y)\sim P}[y\langle\bfw,\phi(\bfx)\rangle]
\end{eqnarray*}
\end{proposition}
{\bf Proof:}
\begin{eqnarray*}
\sup_{\bfw: \|\bfw\| \leq 1} \E_{(\bfx,y)\sim P}[y\langle\bfw,\phi(\bfx)\rangle ] =&& \\
&&\hspace*{-5cm}= \sup_{\bfw: \|\bfw\| \leq 1} \left\langle\bfw,\E_{(\bfx,y)\sim P}[\phi(\bfx) y]\right\rangle \\
&&\hspace*{-5cm}= \left\| \E_{(\bfx,y)\sim P}[\phi(\bfx) y]\right\|\\
&&\hspace*{-5cm}=\sqrt{ \int \int dP(\bfx,y)dP(\bfx',y') \langle \phi(\bfx),\phi(\bfx')\rangle 
yy'}\\
&&\hspace*{-5cm}= \sqrt{\E_{(\bfx,y) \sim P, (\bfx', y')\sim P}[y y' \kappa(\bfx,\bfx')] }
\end{eqnarray*}

The proposition suggests that we can detect useful representations by measuring kernel target alignment. For non-linear functions the difficulty is to identify which combination of features creates a useful representation. We tackle this problem by sampling subsets $S$ of features and assessing whether on average the presence of a particular feature $i$ contributes to an increase $c_i$ in the average kernel target alignment. In this way we derive an empirical estimate of a quantity we will term the contribution.

\begin{definition}
The {\em contribution} $c_i$ of feature $i$ is defined as
\[
c_i = \E_{S \sim \Scal_i}\left[\E_{(\bfx,y) \sim P, (\bfx', y')\sim P}[y y' \kappa_S(\bfx,\bfx')]\right] - 
\E_{S' \sim \Scal_{\setminus i}}\left[\E_{(\bfx,y) \sim P, (\bfx', y')\sim P}[y y' \kappa_{S'}(\bfx,\bfx')]\right],
\]
where $\kappa_S$ denotes the (non-linear) kernel using features in the set $S$ (in our case this will be a Gaussian kernel with equal width), $\Scal_i$ the uniform distribution over sets of features of size $\lfloor n/2\rfloor + 1$ that include the feature $i$, $\Scal_{\setminus i}$ the uniform distribution over sets of features of size $\lfloor n/2 \rfloor$ that do not contain the feature $i$, and $n$ is the number of features. 
\end{definition}

Note that the two distributions over features $\Scal_i$ and $\Scal_{\setminus i}$ are matched in the sense that for each $S$ with non-zero probability in $\Scal_{\setminus i}$, $S \cup \{i\}$ has equal probability in $\Scal_i$. This approach is a straightforward extension of the idea of BaHsic [5].

We will show that for variables that are independent of the target this contribution will be negative. On the other hand, provided there are combinations of variables including the given variable that can generate significant correlations then the contribution of the variable will be positive.

\begin{definition}
We will define an {\em irrelevant} feature to be one whose value is statistically independent of the label and of the other features. 
\end{definition}

We would like an assurance that irrelevant features do not increase alignment. This is guaranteed for the Gaussian kernel by the following result.

\begin{proposition}
Let $P$ be a probability distribution on the product space $\Xcal \times \RR$, where $\Xcal$ has a projection $\phi_Si$ into a Hilbert space $\Fcal$ defined by the Gaussian kernel $\kappa_S$ on a set of features $S$. Suppose a feature $i \not \in S$ is irrelevant.
We have that
\[
\E_{(\bfx,y) \sim P, (\bfx', y')\sim P}[y y' \kappa_{S\cup\{i\}}(\bfx,\bfx')] \leq
\E_{(\bfx,y) \sim P, (\bfx', y')\sim P}[y y' \kappa_{S}(\bfx,\bfx')] 
\]
\end{proposition}
{\bf Proof (sketch):} Since the feature is independent of the target and the other features, functions of these features are also independent. Hence,
\begin{eqnarray*}
\E_{(\bfx,y) \sim P, (\bfx', y')\sim P}[y y' \kappa_{S\cup\{i\}}(\bfx,\bfx')]
&&= \E_{(\bfx,y) \sim P, (\bfx', y')\sim P}[y y' \kappa_{S}(\bfx,\bfx')\exp(- \gamma (x_i - x'_i)^2)]\\
&&\hspace*{-2.5cm}= \E_{(\bfx,y) \sim P, (\bfx', y')\sim P}[y y' \kappa_{S}(\bfx,\bfx')]\E_{(\bfx,y) \sim P, (\bfx', y')\sim P}[\exp(- \gamma (x_i - x'_i)^2)]\\
&&\hspace*{-2.5cm}= \E_{(\bfx,y) \sim P, (\bfx', y')\sim P}[y y' \kappa_{S}(\bfx,\bfx')]\alpha
\end{eqnarray*}
for $\alpha = \E_{(\bfx,y) \sim P, (\bfx', y')\sim P}[\exp(- \gamma (x_i - x'_i)^2)] \leq 1$.

\begin{figure}[h]T
\begin{center}
\includegraphics[width=\textwidth]{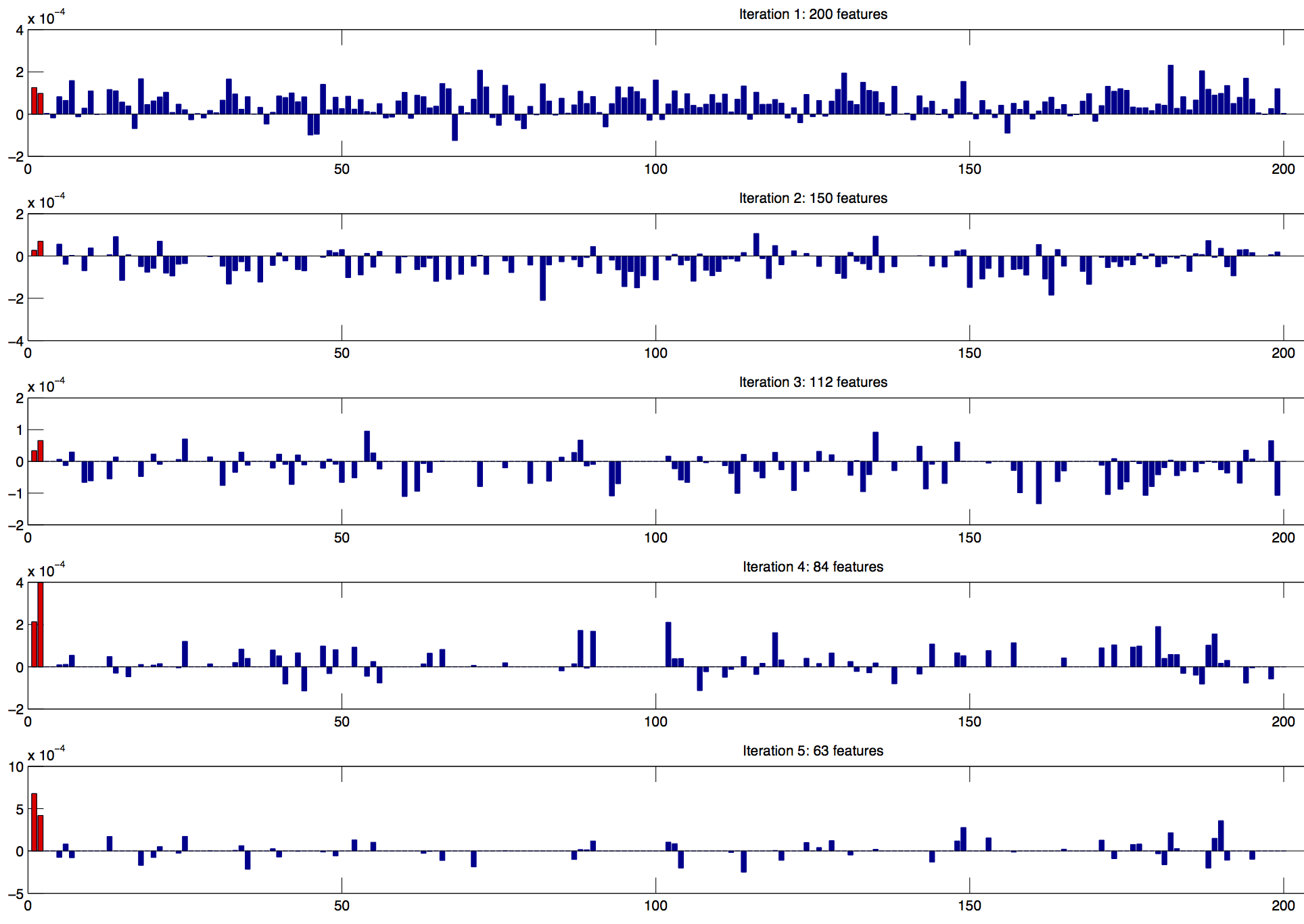}%
\end{center}
\caption{200-dimensional XOR classification problem. The expected contribution of the two relevant features is in red. It can be seen that as more of the noise features are removed in latter iterations of the method, the expected contribution of the two relevant variables rises substantially, in contrast to the contribution of the other features. }
\label{xor}
\end{figure}

In fact the quantity $\alpha$ is typically less than 1 so that adding irrelevant features decreases the alignment. Our approach will be to progressively remove sets of features that are deemed to be irrelevant, hence increasing the alignment together with the signal to noise ratio for the relevant features. Figure \ref{xor} shows how progressively removing features from a learning problem whose output is the XOR function of the first two features both increases the alignment contributions and helps to highlight the two relevant features.

We now introduce our definition of a relevant feature.
\begin{definition}
A feature $i$ will be termed $\eta$-{\em influential} when its contribution $c_i \geq \eta > 0$.
\end{definition} 

So far have only considered expected alignment.  In practice we must estimate this expectation from a finite sample.  We will omit this part of the analysis as it is a straigthforward application of U-statistics that ensures that with high probability for a sufficiently large sample from $\Scal_i$ and $\Scal_{\setminus i}$ and of samples from $P$ (whose sizes depend on $\eta$, $\delta$, the number $k$ of $\eta$-influential variables and the number $T$ of iterations) an empirical estimate of the contribution of an $\eta$-influential variable will with probability at least $1 - \delta$ be greater than 0 for all of the fixed number $T$ of iterations of the algorithm.

Our final ingredient is a method of removing irrelevant features that we will term culling. At each iteration of the algorithm the contributions of all of the features are estimated using the required sample size and the contributions are sorted. We then remove the bottom 25\% of the features in this ordering. Our main result assures us that culling will work under the assumption that the irrelevant variables are independent. 

\begin{theorem}
Fix $\eta> 0$. Suppose that there are $k$ $eta$-influential variables and all other variables are irrelevant. Fix $\delta > 0$ and number $T$ of iterations. Given sufficiently many samples as described above the culling algorithm will with probability at least $1 - \delta$ remove only irrelevant variables will be removed.
\end{theorem}
{\bf Proof (sketch):} Each irrelevant variable has expected contribution less than the contributions of the all the influential features. Hence, with high probability at least 30\% of these features will have lower contributions than all the influential features. Hence, the bottom 25\% will all be irrelevant.

\section{Properties of the algorithm}

We now define our algorithm for randomised selection (randSel). Given a $m \times n$ input matrix $X$ and corresponding output matrix $Y$, randSel proceeds by estimating the individual contribution of features by estimating the alignment of a number of random subsamples that include $n\over 2$ and ${n\over 2}+1$ randomly selected features. This leads to an estimate for the expected alignment contribution of including a feature. The algorithm is parametrized by the number of bootstraps $N$, a bootstrap size$n_b$ and a percentage $z\%$ of features that are dropped after $N$ bootstraps. The algorithm proceeds iteratively until only two features remain. Optionally the algorithm can be further parametrized by permanently including features which were ranked in the top percentile $a\%$ on at least a number $t$ occasions. This option enhances the probability of detecting non-linear dependencies between variables, should they be present. 

There are a number of benefits to this approach, aside from the tangible probabilistic guarantees. RandSel scales gracefully. Considering the computation of a kernel $k(x,x')$ for samples $x, x'$ atomic, the number of kernel computations for a single iteration are $ n_b^2 N$, which for a sensible choice of $N$ can be substantially smaller than the $m^2 n$ complexity of HSIC variants. For example setting $n_b= \sqrt{m}$ and $N=n$ an iteration would require $mn$ kernel element computations, and in addition this process is trivial to parallelize. 

\begin{algorithm}[h!]
   \caption{randSel}
   \label{alg:example}
\begin{algorithmic}
   \STATE {\bfseries Input:} input data $X$, labels $Y$, number of iterations $r$ subsample size $s$, number of features $n$, drop percentile proportion $z$, top percentile proportion $a$, number of occasions $t$
   \REPEAT   
   \FOR{$i=1$ {\bfseries to} $r$}
   \STATE $(X_{i},Y_{i})$ = Random subsample of size $s$ over ${n\over 2}$ randomly selected variables
   \STATE $a_i$= alignment($X_i, Y_i$)
   \STATE $(X^{(+)}_{i},Y^{(+)}_{i})$ = Random subsample of size $s$ over ${n\over 2}+1$ randomly selected variables
   \STATE $a_i^{(+)}$= alignment($X_i^{(+)}, Y_i^{(+)}$)
   \ENDFOR
   \FOR{$j=1$ {\bfseries to} $n$}   
   \STATE mean contribution $c_i$ = mean( $a^{(+)}_i$) - mean($a_i$), where  $j \in X^{(+)}_i$ and $j \notin X_i$
   \ENDFOR
   \STATE drop the  $z$\% bottom-contributing features
   \STATE save the $a$\% top-contributing features
   \IF{ fixing features}
   \IF{ $j$ top-contributor for $t$ consecutive times}
   	\STATE fix feature $j$
     \ENDIF
    \ENDIF 
   \UNTIL{no features left to fix, or only 2 features remain}
   \STATE Return Sequence of estimated contributions and Fixed Variables 
\end{algorithmic}
\end{algorithm}

\section{Results} 

In this section, we present several experiments comparing our feature selection approach to a number of competing methodologies. We have used three synthetic datasets in order to better illustrate the performance characteristics of these algorithms before proceeding to experiments on real data arising in infectious disease profiling. 

\subsection{Experimental Setup}

In both synthetic and real datasets we used nested 10-fold  cross validation to perform feature selection, and repeated the simulations on three different reshuffles of the dataset to account for variance. For every iteration we estimate the validation error after feature selection before proceeding to test on the held out test-set. The inner cross-validation loop determines the number of features to use in classifying the test-set for optimal accuracy. If two or more models are tied in terms of performance, the more parsimonious model is preferred. 

We compare our proposed approach to kernel based algorithms like {\bf RFE}, {\bf FoHsic} and {\bf BaHsic},  as well as a filtering approach relying on {\bf Correlation Coefficients} and {\bf Stability Selection} using the Lasso as the underlying sparse selection algorithm. The same range of gaussian kernel bandwidths was explored in all algorithms and the resulting final classifiers employed  a regularisation parameter of $c=1$.

\subsection{Synthetic Data}

We generated three synthetic datasets in order to carefully illustrate the properties of the different feature selection algorithms. All three synthetic datasets contain 300 samples with a dimensionality of 100 features. The linear and non-linear weston datasets were generated according to [7], and consist of 5 relevant and 95 noise features. Neither the linear or non-linear Weston datasets exhibit a nonlinear interdependence between features. We produced a simple XOR pattern dataset in order to simulate this scenario. Along with the accuracy  on the test set and the sparsity we also record the precision and recall of the selection algorithms. Analogously to information retrieval, we define the precision as the number of the relevant features that were selected from the feature selection procedure over the total number of features selected and recall as the number of relevant features selected over the total number of relevant features.

\begin{table}[h!]
\caption{Results on synthetic data.}
\label{sample-table}
\begin{center}
\begin{tabular}{ l l l l l l l l l|| || }
\hline
\multicolumn{1}{|c}{\bf Dataset}  &\multicolumn{1}{c}{\bf Algorithm  }&\multicolumn{1}{c}{\bf Accuracy  } &\multicolumn{1}{c}{\bf Features} &\multicolumn{1}{c}{\bf Precision} & \multicolumn{1}{c|}{\bf Recall}\\
\hline
Linear Weston  & randSel & 97.7 $\pm$ 2.0 & 3.0 $\pm$ 0.0 & 91.8 $\pm$ 23.1 & 72.0 $\pm$ 16.6  \\
& BaHsic & 97.3 $\pm$ 3.1 & 5.0 $\pm$ 0.0 & 91.5 $\pm$ 19.4 & 70.7 $\pm$ 14.9 \\
& FoHsic & 97.1 $\pm$ 3.1 & 6.0 $\pm$ 0.0 & 95.9 $\pm$ 12.0 & 74.7 $\pm$ 17.7 \\
& Corr. Coeff. & 92.4 $\pm$ 7.8 & 4.0 $\pm$ 0.0 & 96.1 $\pm$ 15.1 & 76.0 $\pm$ 15.5 \\
& Stab. Sel. & 97.3 $\pm$ 3.1 & 2.0 $\pm$ 0.0 & 100.0 $\pm$ 0.0 & 40.0 $\pm$ 0.0 \\
& RFE & 95.3 $\pm$ 3.9 & 5.0 $\pm$ 0.0 & 66.9 $\pm$ 33.7 & 56.0 $\pm$ 13.5 \\
\hline
Non-Linear Weston  & randSel & 99.0 $\pm$ 1.4 & 5.0 $\pm$ 0.0 & 100.0 $\pm$ 0.0 & 89.3 $\pm$ 12.8 \\
& BaHsic & 99.8 $\pm$ 0.9 & 4.0 $\pm$ 0.0 & 100.0 $\pm$ 0.0 & 80.0 $\pm$ 7.6 \\
& FoHsic & 99.8 $\pm$ 0.9 & 4.0 $\pm$ 0.0 & 100.0 $\pm$ 0.0 & 82.7 $\pm$ 7.0 \\
& Corr. Coeff. & 56.2 $\pm$ 6.8 & 21.0 $\pm$ 0.0 & 1.7 $\pm$ 2.5 & 18.7 $\pm$ 31.6 \\
& Stab. Sel. & 50.0 $\pm$ 7.1 & 2.0 $\pm$ 0.0 & 0.0 $\pm$ 0.0 & 0.0 $\pm$ 0.0 \\
& RFE & 98.9 $\pm$ 2.7 & 5.0 $\pm$ 0.0 & 97.8 $\pm$ 5.9 & 100.0 $\pm$ 0.0 \\

\hline
XOR & randSel & 95.7 $\pm$ 3.3 & 2.0 $\pm$ 0.0 & 100.0 $\pm$ 0.0 & 100.0 $\pm$ 0.0 \\
& BaHsic & 95.7 $\pm$ 3.3 & 2.0 $\pm$ 0.0 & 100.0 $\pm$ 0.0 & 100.0 $\pm$ 0.0 \\
& FoHsic & 52.0 $\pm$ 6.5 & 53.0 $\pm$ 0.0 & 9.4 $\pm$ 25.3 & 36.7 $\pm$ 44.2 \\
& Corr. Coeff. & 58.1 $\pm$ 14.9 & 8.0 $\pm$ 0.0 & 10.4 $\pm$ 10.3 & 50.0 $\pm$ 42.3 \\
& Stab. Sel. & 49.3 $\pm$ 11.1 & 2.0 $\pm$ 0.0 & 13.3 $\pm$ 22.9 & 13.3 $\pm$ 22.9 \\
& RFE & 91.8 $\pm$ 12.1 & 2.0 $\pm$ 0.0 & 96.7 $\pm$ 12.9 & 96.7 $\pm$ 12.9 \\
\hline
\end{tabular}
\end{center}
\end{table}

For the synthetic benchmarks we used randSel using 3000 bootstraps of size $m/4$ of the dataset, culling the bottom $25\%$ of variables in terms of expected contribution after the end of each iteration. We did not employ fixing of variables, and the algorithm would iterate until only two variables remained. In the Linear dataset all methods perform fairly well in terms of accuracy with randSel being marginally better. Stability selection is the only method that consistently selects only relevant features, while correlation filtering is marginally better in terms of recall. As is to be expected owing to their linear nature, both correlation filtering and stability selection fail on the Non-linear Weston benchmark. BahSic and FohSic perform better in terms of accuracy, achieving a nearly identical performance on measured variables, while RFE is the only method to achieve perfect recall throughout all folds of the data. Finally in the XOR problem randomised selection and BahSic achieve identical performance across the board. The greedy forward selection employed in FohSic completely fails to identify the two relevant features.

\subsection{Real Data}

We conducted experiments in real datasets arising in the computational profiling of tuberculosis (TB), an application where feature selection  plays a pivotal role both in terms of improving accuracy but also providing insight into the underlying mechanisms. We conducted experiments on two different datasets. The first TB dataset consists of $523$-dimensional mass-spectrometry proteomic profiles of blood plasma [8], and consists of 100 active TB samples, 40 symptomatic controls, and 49 samples of patients with TB-Like symptoms with a co-existing latent TB infection (LTBI).  We performed pairwise comparisons between active TB and Unhealthy Controls \emph{(Task 1)}, Active TB and symptomatic LTBI \emph{(Task 2)}, and Active TB with symptomatic patients without LTBI \emph{(Task 3)},which correspond to scenarios in real clinical applications. The second dataset comprises of the transcriptomic profiles of 69 healthy individuals with LTBI and 133 healthy controls from [9]. Preprocessing removed probes with low acquisition precision as well as factors with missing values, resulting in a set of 6247 variables. Table 2 summarises the experiments.

\begin{table}[h!]
\caption{Results on real data.}
\label{sample-table}
\begin{center}
\begin{tabular}{ l l l l l l l l l|| || }
\hline
\multicolumn{1}{|c}{\bf Dataset}  &\multicolumn{1}{c}{\bf Algorithm  }&\multicolumn{1}{c}{\bf Accuracy  } &\multicolumn{1}{c|}{\bf Features} \\
\hline
TB Task 1  & randSel & 82.9 $\pm$ 8.4 & 64.6 $\pm$ 70.3 \\ 
& BaHsic & 81.7 $\pm$ 9.0 & 74.7 $\pm$ 101.3 \\ 
& FoHsic & 81.3 $\pm$ 9.4 & 68.0 $\pm$ 66.5 \\ 
& Corr. Coeff. & 82.4 $\pm$ 8.8 & 123.6 $\pm$ 85.8 \\ 
& Stab. Sel. & 82.9 $\pm$ 7.3 & 121.7 $\pm$ 56.4 \\ 
& RFE & 81.9 $\pm$ 8.0 & 236.2 $\pm$ 160.2 \\ 
\hline
TB Task 2 & randSel & 82.0 $\pm$ 8.6 & 42.0 $\pm$ 47.7 \\ 
& BaHsic & 81.1 $\pm$ 8.9 & 33.1 $\pm$ 40.6 \\ 
& FoHsic & 80.6 $\pm$ 10.8 & 31.1 $\pm$ 35.3 \\ 
& Corr. Coeff. & 82.7 $\pm$ 9.4 & 73.4 $\pm$ 55.5 \\ 
& Stab. Sel. & 80.7 $\pm$ 8.4 & 137.3 $\pm$ 154.7 \\ 
& RFE & 80.2 $\pm$ 9.1 & 82.4 $\pm$ 139.9 \\ 
\hline
TB Task 3 & randSel & 86.0 $\pm$ 8.1 & 45.3 $\pm$ 33.6 \\ 
& BaHsic & 85.6 $\pm$ 9.5 & 53.3 $\pm$ 39.5 \\ 
& FoHsic & 85.6 $\pm$ 8.8 & 53.6 $\pm$ 44.7 \\ 
& Corr. Coeff. & 85.4 $\pm$ 8.8 & 132.9 $\pm$ 89.7 \\ 
& Stab. Sel. & 84.1 $\pm$ 9.6 & 60.0 $\pm$ 47.9 \\ 
& RFE & 83.9 $\pm$ 9.2 & 43.5 $\pm$ 71.6 \\ 
\hline
TB Micro Array& randSel & 87.6 $\pm$ 4.9 & 58.5 $\pm$ 93.8 \\ 
& BaHsic & 86.1 $\pm$ 6.4 & 61.2 $\pm$ 94.7 \\ 
& FoHsic & 85.2 $\pm$ 7.9 & 52.5 $\pm$ 92.9 \\ 
& Corr. Coeff. & 84.1 $\pm$ 6.6 & 143.5 $\pm$ 114.2 \\ 
& Stab. Sel. & 87.1 $\pm$ 5.9 & 161.8 $\pm$ 136.0 \\ 
& RFE & 85.7 $\pm$ 6.8 & 158.0 $\pm$ 137.6 \\

\end{tabular}
\end{center}
\end{table}

For the mass spectrometry tasks we used randSel using 5000 bootstraps of size $m/3$ of the dataset, culling the bottom $25\%$ of variables in terms of expected contribution after the end of each iteration. We used similar parameters for the Micro-array dataset but with an increased number of bootstraps of 10000 in order to account for the substantially higher dimensionality of the data. Again, no variables were fixed and the algorithm iterated until only two variables remained. In \emph{Task 1}, Randomized selection is tied with stability selection in terms of accuracy, however on average the randomised recovered feature set is significantly sparser. Interestingly, simple filtering based on correlation coefficients performs strongly in the Mass Spectrometry tasks, often beating the HSIC Variants and in fact gives the highest accuracy for \emph{Task 2}. The only test where all the HSIC variants outperformed correlation filtering is the MicroArray task, which has a substantially increased dimensionality in comparison to the mass spectrometry datasets. 

The results indicate that the HSIC-based variants( randSel,FoHsic \& BaHsic) often recover sparser solutions compared to competing algorithms. Given their mutual reliance on HSIC optimisation, the fact that randSel outperforms the other methods in terms of accuracy can be surprising at first glance. It is instructive at this point to acknowledge that these methods rely on heuristics to solve an NP-hard problem. The synthetic XOR dataset already underlines one scenario where randSel outperforms forward greedy selection. The results on real data, combined with our theoretical guarantees, suggest the possibility of arriving at an improved global solution through randSel's incorporation of stochastic information, in opposition to the strategy of obliviously eliminating the locally optimal variable employed in BaHsic.

\subsection{Learning Deep Representations with LPBoostMKL}
A final experimental application where we employed randomised selection was in the recent Black Box Learning challenge [10][13]. After performing an initial unsupervised feature learning step on the original dataset using Sparse Filtering [11], we performed randomised selection in the resulting representation, creating kernels corresponding to the remaining features after each iteration of the feature selection algorithm. Treating each kernel as defining a class of weak learners, we used LPBoost to perform multiple kernel learning (LPBoostMKL [12]). The resulting classifier beat many of our other approaches and was one of the strongest performers in the challenge.

\section{Conclussions}

In this paper we propose randSel, a new algorithm for non-linear feature selection based on randomised estimates of HSIC. RandSel, stochastically estimates the expected importance of features at each iteration, proceeding to cull features uninformative features at the end of each iteration. Our theoretical analysis gives strong guarantees for the expected performance of this procedure which is further demonstrated by testing on a number of real and artificial datasets. This, combined with the algorithm's attractive scaling properties make randSel a strong proposition for use in application areas such as quantitative biology, where the volume of data increases at a frantic pace.

\end{document}